\documentclass[10pt,twocolumn,letterpaper]{article}
\PassOptionsToPackage{table}{xcolor}
\usepackage[pagenumbers]{cvpr}

\usepackage{amsmath,amsfonts,bm}

\def\eqref#1{equation~\ref{#1}}

\def\1{\bm{1}}

\DeclareMathAlphabet{\mathsfit}{\encodingdefault}{\sfdefault}{m}{sl}
\SetMathAlphabet{\mathsfit}{bold}{\encodingdefault}{\sfdefault}{bx}{n}

\usepackage{multirow,makecell}
\definecolor{first}{HTML}{F3B0AE}
\definecolor{second}{HTML}{F8CC98}
\definecolor{third}{HTML}{F8EDC8}

\definecolor{cvprblue}{rgb}{0.21,0.49,0.74}
\usepackage[hypertexnames=false,pagebackref,breaklinks,colorlinks,allcolors=cvprblue]{hyperref}
\hypersetup{pdfauthor={Sicheng Yu, Liu Liu, Xinjie Wang, Xinjiang Wang, Taojun Ding, Wenhao Hu, Beizhen Zhao, Puxin Yan, Zhizhong Su, Hao Wang},pdftitle={GSExplorer: Interactive World Generation with a Persistent 3DGS World State}}
\def\confName{CVPR}
\def\confYear{2027}
\title{GSExplorer: Interactive World Generation with a Persistent 3DGS World State}
\author{
\textbf{Sicheng Yu}$^{1,2}$ \quad
\textbf{Liu Liu}$^{2,\dagger}$ \quad
\textbf{Xinjie Wang}$^{2}$ \quad
\textbf{Xinjiang Wang}$^{2}$ \quad
\textbf{Taojun Ding}$^{2}$\\
\textbf{Wenhao Hu}$^{2}$ \quad
\textbf{Beizhen Zhao}$^{1}$ \quad
\textbf{Puxin Yan}$^{3}$ \quad
\textbf{Zhizhong Su}$^{2}$ \quad
\textbf{Hao Wang}$^{1,\ddagger}$\\[0.5em]
\textnormal{$^{1}$HKUST (GZ) \qquad $^{2}$Horizon Robotics \qquad
$^{3}$The University of Hong Kong}
}

\begin{document}
\maketitle
\begingroup
\renewcommand{\thefootnote}{}
\footnotetext{\hspace*{-1.8em}$^{\dagger}$Project Leader.\quad $^{\ddagger}$Corresponding Author.}
\endgroup
\begin{abstract}
Interactive world generation requires both accurate local control and stable scene consistency over long rollouts. Existing methods improve local generation through camera or geometry conditioning and extend historical context with memory mechanisms. However, long-horizon exploration remains challenging due to incomplete target-view guidance and inconsistencies accumulated during scene updates. We present \textbf{GSExplorer}, a generation--reconstruction framework built around a persistent 3D Gaussian world state. At each step, the current 3DGS scene provides multimodal rendering conditions for a video generator, while camera geometry is injected through a separate Relative Ray Encoding branch. Generated videos are then fed back to a SLAM-based reconstruction backend, where appearance and geometric consistency checks produce a shared support mask for selective scene updates and cross-chunk pose refinement. We further design a data-construction pipeline that reproduces realistic 3DGS degradation during scene extrapolation, yielding 293K training pairs of degraded multimodal renderings and real target videos. GSExplorer achieves the best overall performance among the compared methods on both single-step and three-step WorldScore evaluation, including a Camera Control score of 99.07 in single-step extrapolation. On 481-frame leave-and-return rollouts, it also shows strong revisit consistency.
\iftoggle{cvprfinal}{Project page: \href{https://3dagentworld.github.io/GSExplorer/}{GSExplorer}}{}
\end{abstract}

\section{Introduction}

Starting from a single image, interactive world generation aims to continuously create explorable scenes or videos following user-specified camera trajectories and text prompts~\citep{yu2024wonderjourney,yu2025wonderworld,duan2025worldscore}. This requires two key capabilities: (1) precise camera control with high visual and spatial consistency for each local generation, and (2) stable scene evolution over long rollouts, including revisiting previously explored regions. The former enables reliable scene expansion, while the latter determines whether such capability can persist over continuous exploration.


Existing methods improve local controllability by injecting camera poses, rays, or relative positional encodings into video generators~\citep{he2024cameractrl,dai2026fantasyworld,zhang2026unified-UCPE}, but these signals provide limited information about the scene structure at the target view. Geometry-guided approaches~\citep{yu2024viewcrafter,ren2025gen3c,cao2025uni3c,huang2025voyager,zhou2026uniworld-view,yang2026neoverse,yang2026dreamworld} address this limitation by rendering explicit 3D representations along planned trajectories, providing spatially aligned conditions for generation. Meanwhile, memory-based methods extend temporal context through historical frames, compressed states, or retrieved references to improve long-term consistency~\citep{yu2025context-as-memory,li2025vmem,worldplay2025,team2026dreamx-world}. Recent work further maintains a compact planar memory through geometric merging~\citep{he2026ppmem}. However, long-horizon generation remains challenging: incomplete 3D renderings provide imperfect guidance~\citep{wu2025genfusion,zhang2025scene-splatter}, while pose misalignment and conflicting generated content can accumulate as the scene is repeatedly updated~\citep{zhao2026spatia,wang2026latent-LSM-World}. This motivates a more fundamental question: \textbf{how can we maintain a reliable and queryable 3D world state that continuously provides useful target-view conditions throughout exploration?}

To this end, we introduce \textbf{GSExplorer}, a generation--reconstruction loop built around a persistent 3D Gaussian world state. Our key insight is to separate long-term spatial memory from short-horizon video generation: the persistent 3D scene carries accumulated spatial information, while the video model repeatedly solves local generation problems conditioned on the current world state. Generated videos are selectively integrated back into the same scene, allowing the world state to evolve together with exploration.


Specifically, GSExplorer renders degraded RGB, depth, accumulated opacity, and hole masks from the current 3DGS state as multimodal scene conditions, while camera geometry is injected separately through Relative Ray Encoding to provide complementary cross-view guidance. The generated videos are then fed into a SLAM-based 3DGS reconstruction backend. Inconsistency filtering combines appearance and geometric checks into a support mask for selective scene updates, while cross-chunk pose refinement uses the same mask to filter correspondences and align current and historical views. To train the generator on realistic incomplete scene observations, we construct 293K degraded-condition--target video pairs from real-world scene datasets.

GSExplorer achieves the best overall performance among the compared methods on both single-step and three-step WorldScore evaluation~\citep{duan2025worldscore}. In single-step extrapolation, it achieves a Camera Control score of 99.07 while also attaining leading performance in 3D and Style Consistency. We further evaluate long-horizon revisitation on 50 leave-and-return cases from R2M-Bench~\citep{gu2026r2mbench} with 481-frame rollouts, where GSExplorer achieves the best results among the compared methods across PSNR, SSIM, LPIPS, and Match Ratio. These results demonstrate strong controllability and consistency across both local generation and long-horizon exploration.

Our main contributions are summarized as follows:
\begin{enumerate}
    \item We introduce a generation--reconstruction framework centered on a persistent 3DGS world state, where inconsistency filtering and cross-chunk bundle adjustment enable selective and spatially coherent scene updates across generation steps.

    \item We develop a condition-driven video generator that combines multimodal 3DGS renderings with a separate camera-geometry branch, and construct 293K degraded-condition--target video pairs for accurate camera control and scene completion.
    
    \item GSExplorer achieves leading overall performance on single-step and three-step WorldScore evaluation and shows strong long-horizon revisit consistency after 481-frame rollouts.
\end{enumerate}

\section{Related Work}
\label{sec:related_work}

\subsection{Camera-Controllable World Generation}

Camera-controllable video models encode poses~\citep{wang2024motionctrl}, Pl\"ucker rays~\citep{he2024cameractrl,dai2026fantasyworld}, or relative positions~\citep{li2026cameras-PRoPE,zhang2026unified-UCPE} to guide viewpoint changes. These signals specify camera motion but provide limited target-view scene information. ViewCrafter~\citep{yu2024viewcrafter} guides video diffusion with point-cloud renderings and progressively expands its point cloud through iterative view synthesis. Subsequent methods similarly use rendered geometry for camera control~\citep{ren2025gen3c,cao2025uni3c}.

To handle incomplete renderings, GenFusion~\citep{wu2025genfusion} couples RGB-D restoration with cyclic reconstruction, while Scene Splatter~\citep{zhang2025scene-splatter} uses cascaded momentum to guide video generation and iteratively refine a global Gaussian scene. UniWorld-View~\citep{zhou2026uniworld-view} addresses visibility ambiguity, NeoVerse~\citep{yang2026neoverse} models reconstruction degradation, and DreamWorld~\citep{yang2026dreamworld} completes geometry-aware features before appearance synthesis. However, rendered guidance remains incomplete under large viewpoint changes, making it difficult to preserve scene content while following the intended camera motion. GSExplorer combines complementary multimodal 3DGS renderings with a separate camera-geometry encoding for accurate camera control and robust scene completion.

\subsection{Memory for World Generation}

Memory mechanisms preserve historical information beyond the active context to support long-horizon generation~\citep{chen2026survey}. Context as Memory~\citep{yu2025context-as-memory} retrieves historical frames, while VMem~\citep{li2025vmem} indexes reference views through reconstructed geometry. HY-WorldPlay~\citep{worldplay2025} and DreamX-World~\citep{team2026dreamx-world} likewise reuse historical context during autoregressive generation. Explicit scene representations also support progressive world expansion~\citep{chung2023luciddreamer,yu2024wonderjourney,yu2025wonderworld}. Voyager~\citep{huang2025voyager} and Spatia~\citep{zhao2026spatia} maintain updatable point-cloud memories, while SPMem~\citep{wu2026spmem} combines TSDF-fused static geometry, recent context, and sparse historical frames. LSM-World~\citep{wang2026latent-LSM-World} instead stores diffusion features in a persistent 3D cache.

More recently, PPMem~\citep{he2026ppmem} maintains compact planar memory through geometric merging to guide subsequent generation. However, merging geometry at prescribed camera poses can retain spatial misalignment when generated motion deviates from the intended trajectory. In contrast, GSExplorer maintains a persistent 3DGS world state by estimating camera poses from generated videos and applying inconsistency filtering and support-guided cross-chunk BA, allowing the updated scene to provide reliable target-view conditions for subsequent short-horizon generation.

\begin{figure*}[!t]
    \centering
    \includegraphics[width=0.95\linewidth]{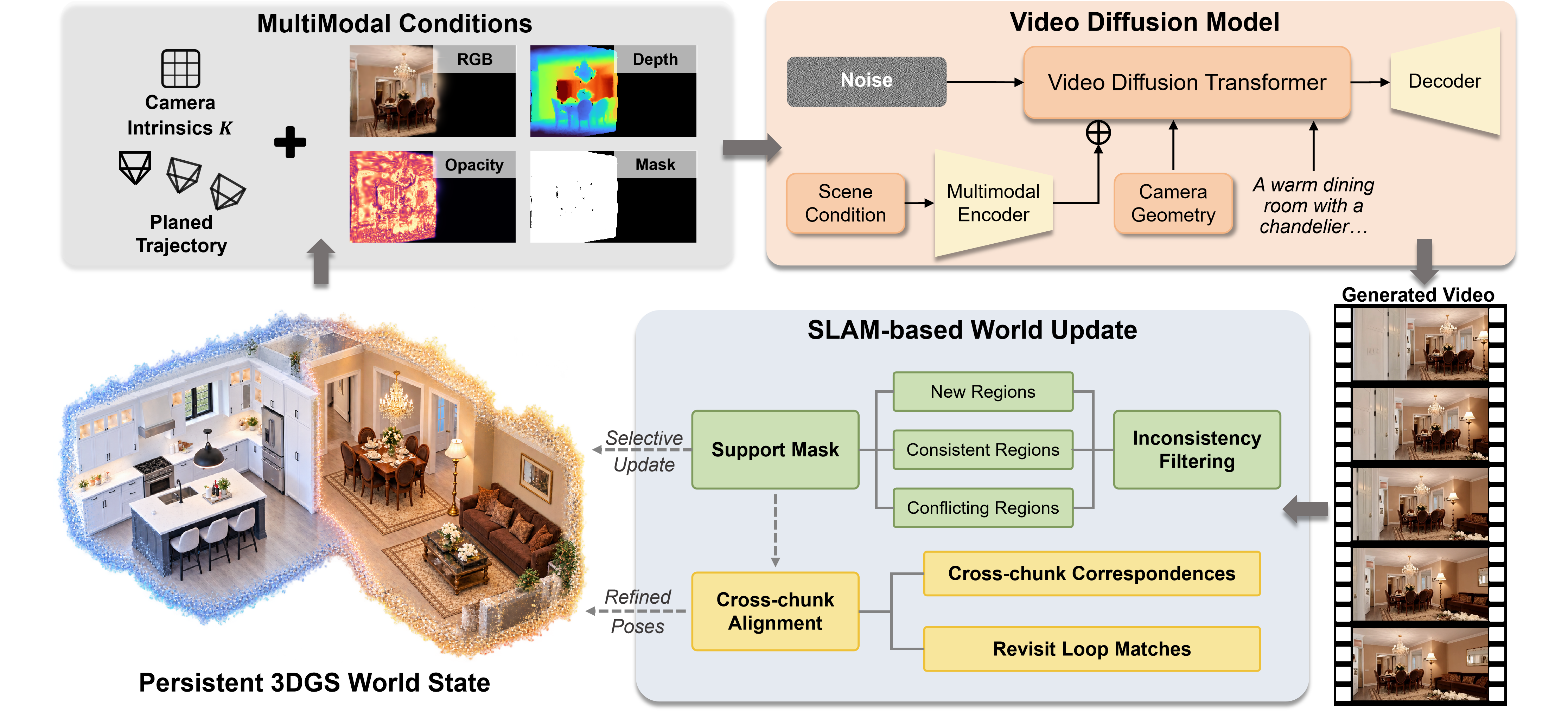}

    \caption{
    \textbf{Overview of GSExplorer.}
    Starting from a 3DGS scene initialized from a single image, we render multimodal scene conditions along the planned camera trajectory at each step.
    These conditions, together with camera geometry and text prompts, guide the condition-driven generator to produce a completed video.
    The generated video is then fed back to the SLAM-based world update, where content and geometric masks are combined into a support mask for selective scene updates.
    The support mask also filters correspondences for cross-chunk alignment, as indicated by the dashed arrow.
    The updated 3DGS world state then provides reliable conditions for the next step, forming a generation--reconstruction loop.
    }

    \vspace{-5pt}
    
    \label{fig:3.1-pipeline}
\end{figure*}

\section{METHODOLOGY}
Figure~\ref{fig:3.1-pipeline} gives an overview of GSExplorer. 
We present the persistent 3DGS world update in Sec.~\ref{sec3-1:world_state}, the condition-driven generation model in Sec.~\ref{3.2-generation model}, and the training-pair construction in Sec.~\ref{sec3-3:training_pairs}.

\subsection{Persistent 3DGS World State}
\label{sec3-1:world_state}

A persistent world state should preserve existing information and remain reliable and queryable as newly generated videos are incorporated. Generated videos may reveal new regions, provide consistent views of existing areas, or introduce conflicting content. Incorporating all observations indiscriminately can rapidly accumulate errors. We maintain a 3DGS scene with a SLAM-based reconstruction backend and compute a support mask to maintain consistency during pose alignment and scene updates.

\paragraph{Inconsistency filtering and selective update.}
For each generated frame $I_i$, we render the historical Gaussian scene at its estimated camera pose to obtain the rendered image $\hat{I}_i^{\mathrm{old}}$ and accumulated opacity $A_i^{\mathrm{old}}$. We identify sufficiently covered regions through opacity:
\begin{equation}
    H_i(u)
    =
    \mathbf{1}\!\left[A_i^{\mathrm{old}}(u)\geq\eta\right].
    \label{eq:historical_coverage}
\end{equation}
Regions with $H_i(u)=0$ are treated as new regions, where new Gaussians can be inserted and optimized.

In covered regions that overlap with historical frames, we preserve the existing scene when new observations conflict with it. 
We first filter appearance conflicts by comparing the generated frame with the historical rendering:
\begin{equation}
    e_i^{\mathrm{app}}(u)
    =
    \frac{1}{3}
    \left\|I_i(u)-\hat{I}_i^{\mathrm{old}}(u)\right\|_1.
    \label{eq:appearance_discrepancy}
\end{equation}
The content mask is $C_i(u)=\mathbf{1}[e_i^{\mathrm{app}}(u)\leq\tau_a]$.

\begin{figure*}[!t]
    \centering
    \includegraphics[width=0.98\linewidth]{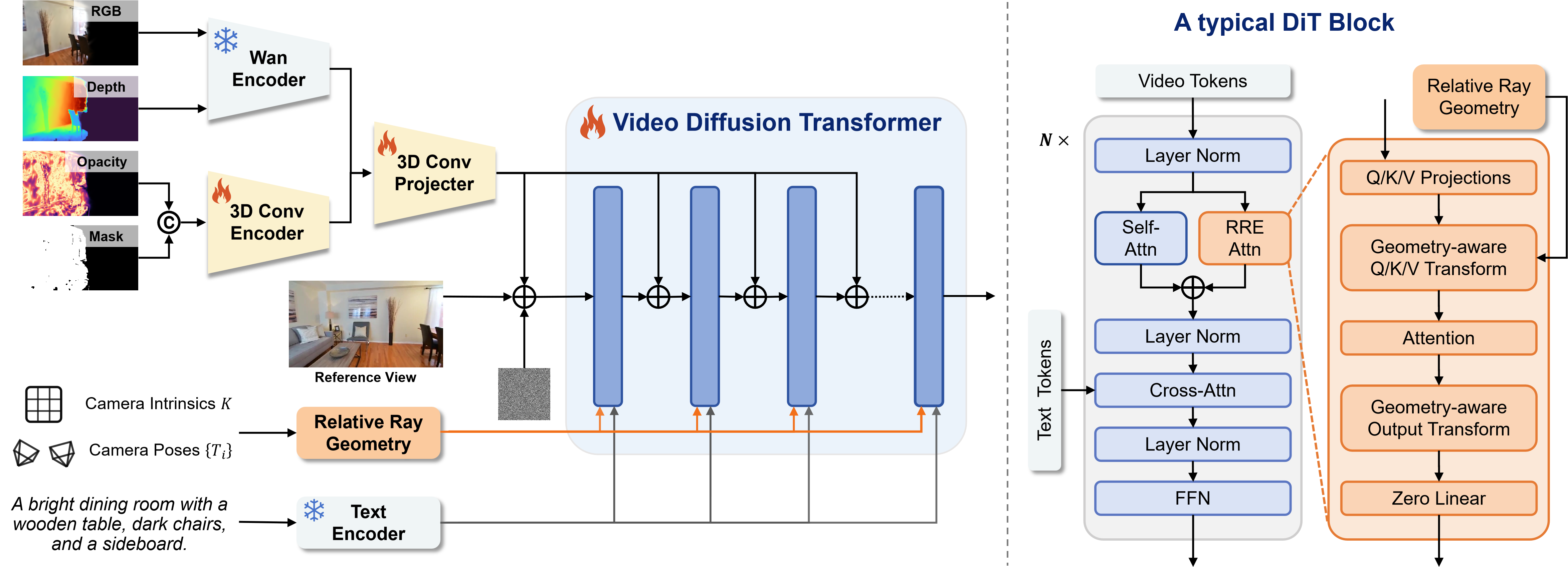}

    \vspace{-2pt}
    
    \caption{
    \textbf{Architecture of the condition-driven video generator.}
    Multimodal 3DGS renderings are encoded as scene conditions and injected into the diffusion backbone through one input projection and five residual projections.
    Camera intrinsics and poses define token-wise ray geometry, which is incorporated through a separate RRE attention branch in every DiT block using geometry-aware Q/K/V and output transforms.
    Text prompts follow the original Wan cross-attention pathway.}

    \vspace{-6pt}

    \label{fig:3.2-diffusion}
\end{figure*}

We further check geometric consistency using matches among new frames and between new and historical frames. We reuse the SLAM backend's keyframe associations and retrieval results, selecting at most $K$ reference frames per generated frame to bound the cost of this check. For a pixel $u$ matched to $v_{ij}(u)$ in reference frame $j$, we back-project $u$ using its estimated depth $D_i(u)$ and reproject it into frame $j$ using the estimated camera poses. Denoting this projected location by $\hat{x}_{ij}(u)$, the reprojection error is:
\begin{equation}
    e_{ij}^{\mathrm{geo}}(u)
    =
    \left\|\hat{x}_{ij}(u)-x_j(v_{ij}(u))\right\|_2,
    \label{eq:reprojection_error}
\end{equation}
where both coordinates are normalized by camera intrinsics. We aggregate valid matches across the selected reference frames to obtain the geometric mask:
\begin{equation}
    G_i(u)
    =
    \mathbf{1}\!\left[
        \operatorname*{median}_{j\in\mathcal{N}_i(u)}
        e_{ij}^{\mathrm{geo}}(u)
        \leq\tau_g
    \right],
    \label{eq:geometric_mask}
\end{equation}
where $\mathcal{N}_i(u)$ contains reference frames with valid matches for $u$. We set $G_i(u)=1$ when $\mathcal{N}_i(u)=\varnothing$.

The final support mask is:
\begin{equation}
    S_i(u)
    =
    G_i(u)\bigl[(1-H_i(u))+H_i(u)C_i(u)\bigr].
    \label{eq:reconstruction_support}
\end{equation}
Only supported pixels contribute to Gaussian insertion and scene optimization, excluding conflicting observations from scene updates. Prior work corrects inconsistencies in generated videos through non-rigid alignment for 3DGS reconstruction~\citep{hollein2026video-to-world}, but this introduces substantial optimization cost. Our filtering builds on the local consistency of our video generator, aiming to limit error accumulation by discarding a small fraction of observations. 


\paragraph{Support-guided cross-chunk alignment.}
Generated videos may slightly deviate from the planned trajectory, so we estimate poses from the generated sequence. To maintain cross-chunk continuity and revisit consistency, we use the SLAM backend's inter-frame correspondences and verified loop-closure matches for cross-chunk bundle adjustment. We filter these matches with support masks:
\begin{equation}
    \widetilde{\mathcal{C}}_{ij}
    =
    \left\{
        (u,v)\in\mathcal{C}_{ij}^{\mathrm{SLAM}}
        \mid S_i(u)S_j(v)=1
    \right\},
    \label{eq:support_gated_correspondences}
\end{equation}
where $\mathcal{C}_{ij}^{\mathrm{SLAM}}$ contains matches established by the backend. The retained matches constrain cross-chunk bundle adjustment, followed by current-chunk pose refinement. We then refresh the current-frame support masks at the final poses before updating the Gaussian scene. Further details are provided in the Appendix.

\subsection{Condition-Driven Video Generation}
\label{3.2-generation model}

We build our generator on Wan2.2-TI2V-5B~\citep{wan2025wan}, conditioned on multimodal 3DGS renderings, camera geometry, and text prompts. As shown in Fig.~\ref{fig:3.2-diffusion}, rendered observations provide the main scene guidance, while camera geometry is injected through a separate branch as complementary guidance. 
This avoids conflict and allows the two condition streams to cooperate during generation.

\paragraph{Multimodal conditioning.}
Given a planned camera trajectory, we render degraded RGB, depth, and accumulated opacity from the current 3DGS state, and derive a binary hole mask from the opacity. RGB provides observed appearance, depth provides explicit geometry, accumulated opacity describes continuous rendering coverage, and the mask indicates hole regions to be completed.


RGB and normalized depth are encoded separately with the same pretrained Wan VAE, while opacity and the hole mask are mapped to the same latent grid using a lightweight 3D convolutional encoder. 
Their concatenated features are projected to the DiT hidden space and injected through six zero-initialized projections: one at the input and five independent residual projections after selected DiT blocks.

Unlike prior designs that inject camera cues together with rendered modalities~\citep{yang2026neoverse}, we encode camera geometry through a separate Relative Ray Encoding (RRE) branch from UCPE~\citep{zhang2026unified-UCPE}. In this branch, camera intrinsics and poses define token-wise ray geometry, which is used to transform the queries, keys, and values of a lightweight attention module, followed by a geometry-aware output transform before projection back to the backbone features. The resulting RRE attention output is added in parallel to the original self-attention in every DiT block. 
This lightweight design keeps rendered conditions dominant in scene completion, with camera geometry providing complementary guidance for effective control. Depth and camera translation share a clip-level normalization scale for geometric consistency. 

\begin{figure*}[!t]
    \centering
    \includegraphics[width=\linewidth]{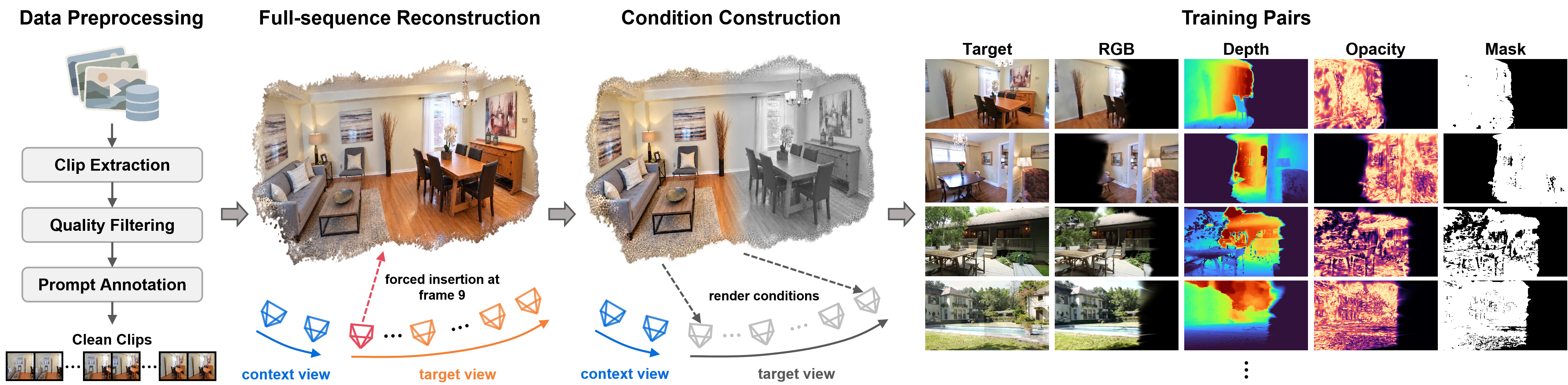}

    \caption{
    \textbf{Training-pair construction pipeline.}
    First, long sequences are filtered and sampled into 57-frame clips containing 8 context frames and 49 target frames.
    Next, we reconstruct the full sequence with 3DGS-SLAM while forcing Gaussian insertion at the first target frame. Together, these steps improve target-view alignment while preserving the desired context-limited degradation.
    We then retain only Gaussians born within the context frames, recovering the desired context-limited degradation.
    Finally, the retained 3DGS is rendered at the target poses to produce multimodal degraded conditions paired with the real target frames.
    }

    \label{fig:3.3-data_pipeline}
\end{figure*}

\begin{table*}[!t]
    \centering
    \caption{\textbf{Results on WorldScore for single-step extrapolation.} GSExplorer achieves the best overall performance with strong controllability and consistency. Avg. is the mean of the seven metrics. The \colorbox{first}{\strut first}, \colorbox{second}{\strut second}, and \colorbox{third}{\strut third} best results are highlighted accordingly.}
    \label{tab:worldscore_single_step}

    \fontsize{8.5}{9.5}\selectfont
    \renewcommand{\arraystretch}{1}
    \setlength{\tabcolsep}{4.5pt}

    \begin{tabular}{l|cccccccc}
        \toprule
        \textbf{Method}
        & \makecell{\textbf{3D}\\\textbf{Consist.}}
        & \makecell{\textbf{Photo}\\\textbf{Consist.}}
        & \makecell{\textbf{Style}\\\textbf{Consist.}}
        & \makecell{\textbf{Camera}\\\textbf{Ctrl.}}
        & \makecell{\textbf{Object}\\\textbf{Ctrl.}}
        & \makecell{\textbf{Content}\\\textbf{Align.}}
        & \makecell{\textbf{Subjective}\\\textbf{Qual.}}
        & \textbf{Avg.} \\
        \midrule
        WonderWorld~\citep{yu2025wonderworld} & 85.55 & 81.70 & 62.36 & 91.77 & 58.69 & \cellcolor{first}75.15 & 44.53 & \textbf{71.39} \\
        GenFusion~\citep{wu2025genfusion} & 86.42 & 78.62 & 58.67 & 88.46 & 57.06 & 44.56 & 44.22 & \textbf{65.43} \\
        AETHER~\citep{zhu2025aether} & 79.84 & 58.68 & 72.09 & 57.44 & 52.26 & 28.06 & 41.11 & \textbf{55.64} \\
        Uni3C~\citep{cao2025uni3c} & 82.77 & 72.91 & 81.30 & 89.29 & 48.56 & 62.14 & 54.76 & \textbf{70.25} \\
        Gen3C~\citep{ren2025gen3c} & 79.34 & 87.95 & \cellcolor{third}91.24 & \cellcolor{third}92.89 & 61.91 & 46.76 & 42.09 & \textbf{71.74} \\
        Voyager~\citep{huang2025voyager} & 56.00 & 80.68 & 72.89 & 45.92 & 57.69 & 48.36 & 44.74 & \textbf{58.04} \\
        FantasyWorld~\citep{dai2026fantasyworld} & 83.17 & 86.97 & 85.34 & 49.47 & 48.31 & 52.61 & 52.58 & \textbf{65.49} \\
        Wan2.2-5B-Camera~\citep{wan2025wan} & 78.19 & 83.07 & 84.16 & 75.83 & 62.59 & 63.12 & 47.46 & \textbf{70.63} \\
        NeoVerse~\citep{yang2026neoverse} & 83.67 & 78.65 & 89.33 & 90.81 & 31.63 & \cellcolor{third}69.02 & \cellcolor{third}63.03 & \textbf{72.31} \\
        Spatia~\citep{zhao2026spatia} & 86.81 & 86.63 & 85.33 & 73.98 & 58.84 & 51.36 & 49.74 & \textbf{70.38} \\
        DreamWorld~\citep{yang2026dreamworld} & 84.96 & \cellcolor{first}92.70 & \cellcolor{second}92.98 & 81.97 & \cellcolor{third}63.11 & 50.37 & 59.17 & \cellcolor{third}\textbf{75.04} \\
        DreamX-World~\citep{team2026dreamx-world} & 77.93 & 64.81 & 86.21 & 73.30 & \cellcolor{second}63.56 & 68.00 & 50.84 & \textbf{69.24} \\
        UniWorld-View~\citep{zhou2026uniworld-view} & \cellcolor{second}89.53 & \cellcolor{second}89.41 & 84.38 & \cellcolor{second}94.57 & 46.09 & 68.51 & \cellcolor{second}63.76 & \cellcolor{second}\textbf{76.61} \\
        PPMem~\citep{he2026ppmem} & \cellcolor{third}87.79 & 81.75 & 87.92 & 84.41 & 58.84 & 47.58 & 53.56 & \textbf{71.69} \\
        \midrule
        \textbf{GSExplorer (Ours)} & \cellcolor{first}90.83 & \cellcolor{third}88.47 & \cellcolor{first}94.40 & \cellcolor{first}99.07 & \cellcolor{first}65.91 & \cellcolor{second}74.70 & \cellcolor{first}66.59 & \cellcolor{first}\textbf{82.85} \\
        \bottomrule
    \end{tabular}
\end{table*}

\paragraph{Training scheme.}

We train the generator on paired videos using a weighted flow-matching objective. Given the target video latent $z_0$ and Gaussian noise $\epsilon\sim\mathcal{N}(0,I)$, we form $z_\sigma=(1-\sigma)z_0+\sigma\epsilon$ and predict the velocity target $\epsilon-z_0$. Following the TI2V formulation, we replace the first temporal latent slice with the clean encoding of the first target frame and exclude it from the loss when the clean reference is retained. Denoting the resulting input by $\bar{z}_\sigma$, we optimize the expectation over $z_0$, $\epsilon$, and $\sigma$:
\begin{equation}
    \mathcal{L}_{\mathrm{FM}}
    =
    \mathbb{E}\bigl[
        w(\sigma)\,
        \operatorname{MSE}_{\neg\mathrm{ref}}\bigl(
            v_{\theta}(\bar{z}_{\sigma},\sigma;\mathcal{C},y),
            \epsilon-z_0
        \bigr)
    \bigr],
    \label{eq:flow_matching}
\end{equation}
where $\mathcal{C}$ denotes the conditions, $y$ is the text prompt, $w(\sigma)$ is a timestep-dependent weight, and $\operatorname{MSE}_{\neg\mathrm{ref}}$ averages over all latent elements except any retained clean-reference slice.

Training follows two stages. We first keep the DiT backbone frozen and optimize only the multimodal adapter and pose branch to establish basic condition following. We then jointly fine-tune the DiT and both conditioning branches. 

\subsection{Training Pair Construction}
\label{sec3-3:training_pairs}

To train our model to complete videos from multimodal conditions and text prompts, we construct degraded conditions paired with target videos from diverse real-world scene datasets, including RealEstate10K~\citep{zhou2018stereo-re10k}, DL3DV~\citep{ling2024dl3dv}, ACID~\citep{liu2021infinite-acid}, and ScanNet~\citep{dai2017scannet}. As shown in Fig.~\ref{fig:3.3-data_pipeline}, the key is to reproduce realistic 3DGS degradation while maintaining pixel alignment with the real targets.

\paragraph{Data preprocessing.}
We segment long sequences into 57-frame clips after quality filtering and use Qwen3.5~\citep{qwen3.5} to annotate a scene description for each clip. The first 8 frames serve as context and the remaining 49 as targets. More details are provided in the Appendix.

\begin{table*}[!t]
    \centering
    \caption{\textbf{Results on WorldScore for three-step extrapolation.} GSExplorer maintains the best overall performance under multi-step generation, with strong camera control and scene consistency. Averaging and color conventions follow Table~\ref{tab:worldscore_single_step}.}
    \label{tab:worldscore_three_step}

    \fontsize{8.5}{9.5}\selectfont
    \renewcommand{\arraystretch}{1}
    \setlength{\tabcolsep}{4.5pt}

    \begin{tabular}{l|cccccccc}
        \toprule
        \textbf{Method}
        & \makecell{\textbf{3D}\\\textbf{Consist.}}
        & \makecell{\textbf{Photo}\\\textbf{Consist.}}
        & \makecell{\textbf{Style}\\\textbf{Consist.}}
        & \makecell{\textbf{Camera}\\\textbf{Ctrl.}}
        & \makecell{\textbf{Object}\\\textbf{Ctrl.}}
        & \makecell{\textbf{Content}\\\textbf{Align.}}
        & \makecell{\textbf{Subjective}\\\textbf{Qual.}}
        & \textbf{Avg.} \\
        \midrule
        WonderWorld~\citep{yu2025wonderworld} & 73.93 & 36.81 & 49.95 & \cellcolor{first}92.54 & 49.88 & \cellcolor{first}64.89 & 48.53 & \textbf{59.50} \\
        GenFusion~\citep{wu2025genfusion} & 72.51 & 73.94 & 62.24 & 74.71 & 44.26 & 31.78 & 40.20 & \textbf{57.09} \\
        AETHER~\citep{zhu2025aether} & 63.97 & 33.11 & 61.99 & 4.43 & 34.78 & 33.69 & 35.09 & \textbf{38.15} \\
        Uni3C~\citep{cao2025uni3c} & 65.81 & 61.41 & 71.18 & 59.01 & 50.25 & 38.58 & 49.81 & \textbf{56.58} \\
        Gen3C~\citep{ren2025gen3c} & 72.34 & \cellcolor{third}82.08 & 79.57 & \cellcolor{third}80.16 & 48.62 & 38.72 & 49.49 & \cellcolor{second}\textbf{64.43} \\
        Voyager~\citep{huang2025voyager} & 17.62 & 10.93 & 60.43 & 5.80 & 49.11 & 44.20 & 49.31 & \textbf{33.91} \\
        FantasyWorld~\citep{dai2026fantasyworld} & 72.31 & 76.55 & 77.94 & 23.51 & 42.62 & \cellcolor{third}50.72 & 47.68 & \textbf{55.90} \\
        Wan2.2-5B-Camera~\citep{wan2025wan} & 64.73 & 80.05 & 71.07 & 49.38 & 51.62 & 39.57 & 34.97 & \textbf{55.91} \\
        NeoVerse~\citep{yang2026neoverse} & 66.48 & 75.68 & \cellcolor{second}83.99 & 65.09 & 26.50 & 41.74 & \cellcolor{second}61.30 & \textbf{60.11} \\
        Spatia~\citep{zhao2026spatia} & 65.95 & 77.33 & 75.17 & 29.85 & 44.88 & 10.58 & 37.95 & \textbf{48.82} \\
        DreamX-World~\citep{team2026dreamx-world} & 56.69 & 37.59 & \cellcolor{third}80.29 & 33.28 & \cellcolor{third}52.63 & 46.27 & 38.80 & \textbf{49.36} \\
        UniWorld-View~\citep{zhou2026uniworld-view} & \cellcolor{second}75.02 & \cellcolor{first}88.16 & 70.45 & 71.21 & 42.13 & 35.12 & \cellcolor{first}66.49 & \cellcolor{third}\textbf{64.08} \\
        PPMem~\citep{he2026ppmem} & \cellcolor{third}74.85 & 69.55 & 66.69 & 62.80 & \cellcolor{second}53.06 & 39.78 & 45.30 & \textbf{58.86} \\
        \midrule
        \textbf{GSExplorer (Ours)} & \cellcolor{first}77.50 & \cellcolor{second}84.51 & \cellcolor{first}84.36 & \cellcolor{second}85.17 & \cellcolor{first}56.25 & \cellcolor{second}52.82 & \cellcolor{third}60.34 & \cellcolor{first}\textbf{71.56} \\
        \bottomrule
    \end{tabular}
\end{table*}

\begin{figure*}[!t]
    \centering
    \includegraphics[width=\linewidth]{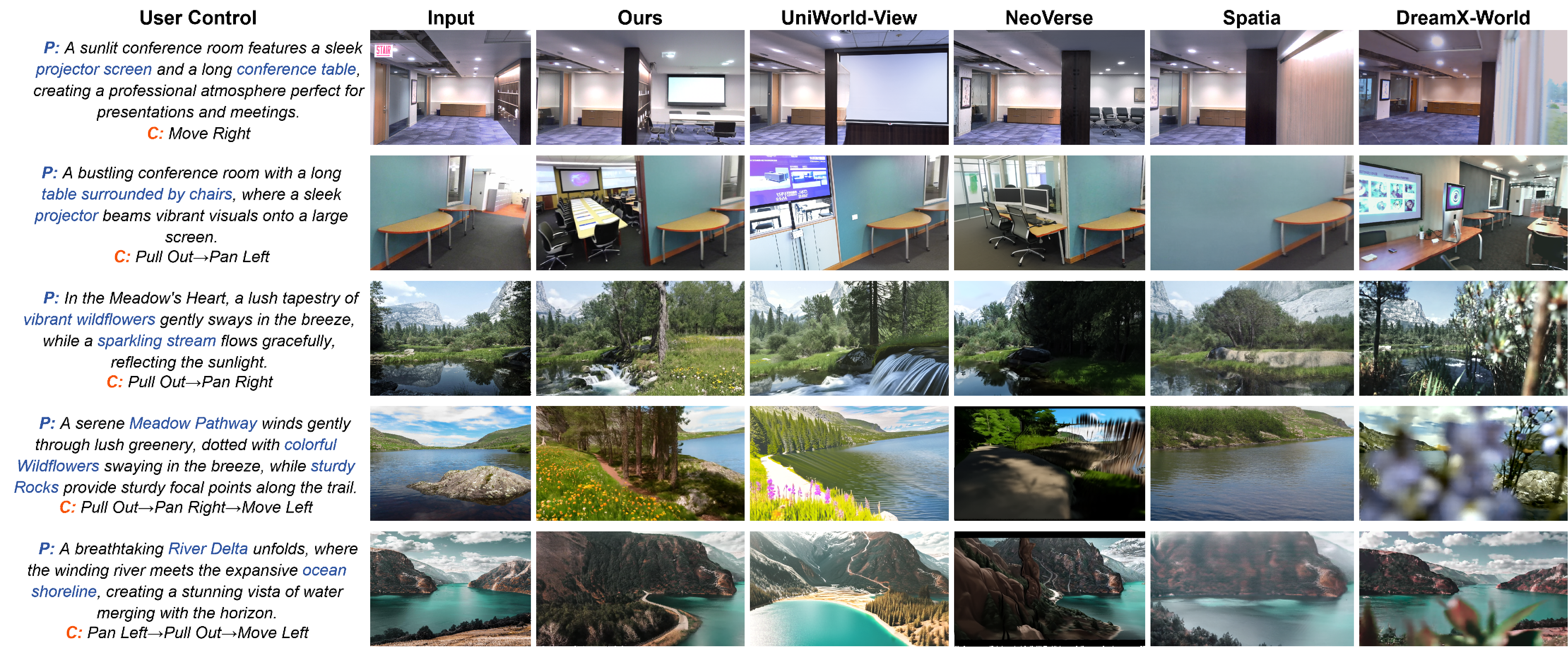}
    \vspace{-6pt}
    \caption{\textbf{Qualitative WorldScore results across one-, two-, and three-step extrapolation.}
    \(P\) and \(C\) denote the text prompt and camera instruction. GSExplorer accurately follows both controls while maintaining high visual quality and coherent spatial structure.}
    \vspace{-5pt}
    \label{fig:main-generation}
\end{figure*}

\paragraph{Degraded condition construction.}
Reconstructing only the 8 context frames provides insufficient geometric constraints, causing rendered conditions to become severely misaligned with real targets at farther extrapolated views. We therefore run the same 3DGS-SLAM system in our extrapolation pipeline over the full 57-frame sequence and record the birth frame \(b_g\) of each Gaussian primitive.


Full-sequence optimization, however, may cause context Gaussians to adapt to future observations and weaken the degradation we aim to simulate. We therefore force the first target frame (index 8, zero-based) to be a keyframe and insert new Gaussians for newly revealed content. After reconstruction, we retain only Gaussians born within the context:
\begin{equation}
    \mathcal{G}^{\mathrm{ctx}}
    =
    \left\{
    g \in \mathcal{G}^{\mathrm{full}}
    \mid b_g < 8
    \right\}.
    \label{eq:context_gaussians}
\end{equation}
The retained Gaussians keep their optimized parameters to preserve alignment with the real targets.

\paragraph{Condition rendering and alignment filtering.}
We render \(\mathcal{G}^{\mathrm{ctx}}\) at the 49 SLAM-estimated target poses to obtain degraded RGB, depth, and accumulated opacity, derive the corresponding hole masks, and retain the camera poses as conditions. To filter remaining misaligned samples, we apply ECC registration~\citep{evangelidis2008parametric-ECC-alignment} between the valid region of the last rendered frame and its target and compute the mean pixel displacement \(d_{\mathrm{ECC}}\). Clips with \(d_{\mathrm{ECC}}>\tau\) are discarded. The final dataset contains 293K video pairs.

\begin{figure*}[!t]
    \centering
    \includegraphics[width=\linewidth]{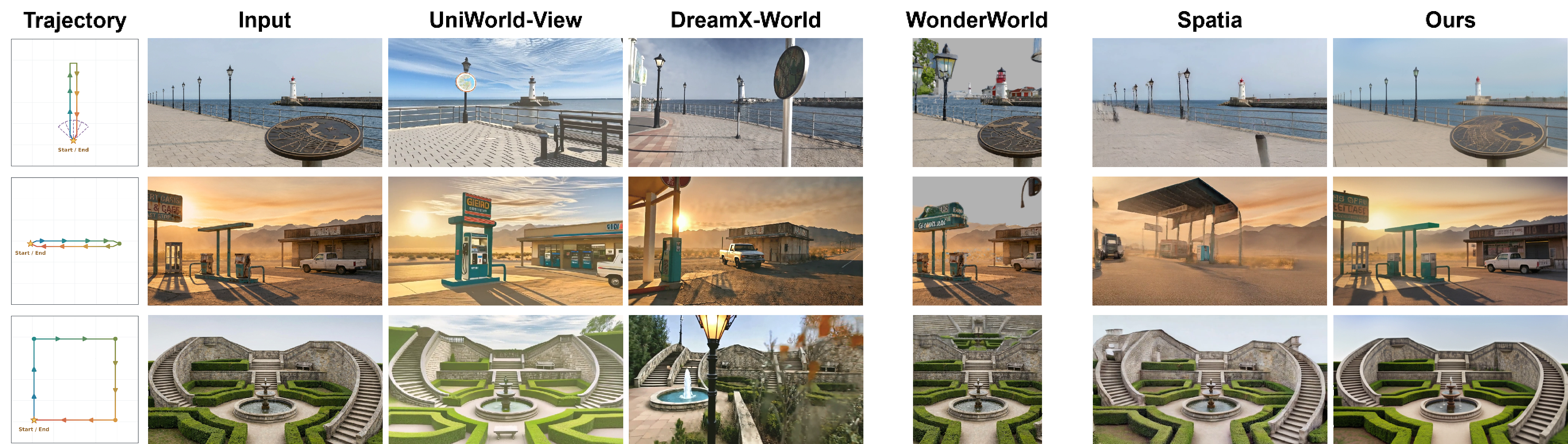}

    \caption{\textbf{Qualitative comparison of long-horizon revisit results.} GSExplorer better preserves the content and details of the starting view after long-range exploration.}

    \vspace{-5pt}
    
    \label{fig:main-revisit}
\end{figure*}

\section{Experiments}

\subsection{Experimental Setup}

\paragraph{Benchmarks and metrics.}
We evaluate both single and three step world extrapolation on the photorealistic subset of the WorldScore static benchmark~\citep{duan2025worldscore}. Following its protocol, we report seven metrics covering controllability and quality: Camera Control, Object Control, Content Alignment, 3D Consistency, Photometric Consistency, Style Consistency, and Subjective Quality. For long-horizon revisit evaluation, we select 50 cases from R2M-Bench~\citep{gu2026r2mbench} and use its leave-and-return trajectories. Each method generates a 10-step, 481-frame rollout that returns to the initial viewpoint. We compare the final frame with the input image using PSNR, SSIM~\citep{wang2004ssim}, LPIPS~\citep{zhang2018unreasonable-lpips}, and Match Ratio. Following R2M-Bench, Match Ratio is computed using SuperPoint~\citep{detone2018superpoint} and LightGlue~\citep{lindenberger2023lightglue}, normalized by the smaller number of detected keypoints in the image pair.

\begin{table}[!t]
    \centering
    \caption{\textbf{Long-horizon revisit evaluation.} GSExplorer achieves the best performance across all four metrics.}
    \label{tab:revisit_eval}
    \vspace{-3pt}
    \fontsize{8.5}{9.5}\selectfont
    \renewcommand{\arraystretch}{1}
    \setlength{\tabcolsep}{4pt}

    \begin{tabular}{lcccc}
        \toprule
        \textbf{Method}
        & \textbf{PSNR}$\uparrow$
        & \textbf{SSIM}$\uparrow$
        & \textbf{LPIPS}$\downarrow$
        & \makecell{\textbf{Match}\\\textbf{Ratio}$\uparrow$} \\
        \midrule
        UniWorld-View~\citep{zhou2026uniworld-view}
        & 11.73 & 0.375 & 0.544 & 0.210 \\
        DreamX-World~\citep{team2026dreamx-world}
        & 12.02 & 0.417 & 0.649 & 0.131 \\
        WonderWorld~\citep{yu2025wonderworld}
        & 15.09 & 0.592 & 0.427 & 0.431 \\
        Spatia~\citep{zhao2026spatia}
        & 17.86 & 0.565 & 0.415 & 0.423 \\
        PPMem~\citep{he2026ppmem}
        & 12.05 & 0.442 & 0.529 & 0.376 \\
        \midrule
        \textbf{GSExplorer (Ours)}
        & \textbf{19.85}
        & \textbf{0.623}
        & \textbf{0.369}
        & \textbf{0.481} \\
        \bottomrule
    \end{tabular}
    \vspace{-3pt}
\end{table}

\paragraph{Baselines.}
WorldScore comparisons include WonderWorld~\citep{yu2025wonderworld}, GenFusion~\citep{wu2025genfusion}, AETHER~\citep{zhu2025aether}, Uni3C~\citep{cao2025uni3c}, Gen3C~\citep{ren2025gen3c}, Voyager~\citep{huang2025voyager}, FantasyWorld~\citep{dai2026fantasyworld}, Wan2.2-5B-Camera~\citep{wan2025wan}, NeoVerse~\citep{yang2026neoverse}, Spatia~\citep{zhao2026spatia}, DreamWorld~\citep{yang2026dreamworld}, DreamX-World~\citep{team2026dreamx-world}, UniWorld-View~\citep{zhou2026uniworld-view}, and PPMem~\citep{he2026ppmem}. These cover explicit 3D scene generation, geometry-guided video generation, and interactive world models. Revisit comparisons include UniWorld-View, DreamX-World, WonderWorld, Spatia, and PPMem.

\paragraph{Implementation details.}
We initialize the generator from Wan2.2-TI2V-5B and train it for 30K iterations on 32 NVIDIA H20 GPUs with a global batch size of 32. Training clips contain 49 frames at $640\times352$ resolution, while ScanNet clips use $640\times480$. For the first 10K iterations, we train the multimodal adapter and pose branch with a learning rate of $10^{-4}$. We then jointly fine-tune the DiT, adapter, and pose branch for 20K iterations with learning rates of $1\times10^{-7}$, $1\times10^{-5}$, and $1\times10^{-5}$, respectively. 
All evaluations are performed on a single NVIDIA H20 GPU.
Additional training details are provided in the Appendix.

\begin{figure}[!t]
    \centering
    \includegraphics[width=\linewidth]{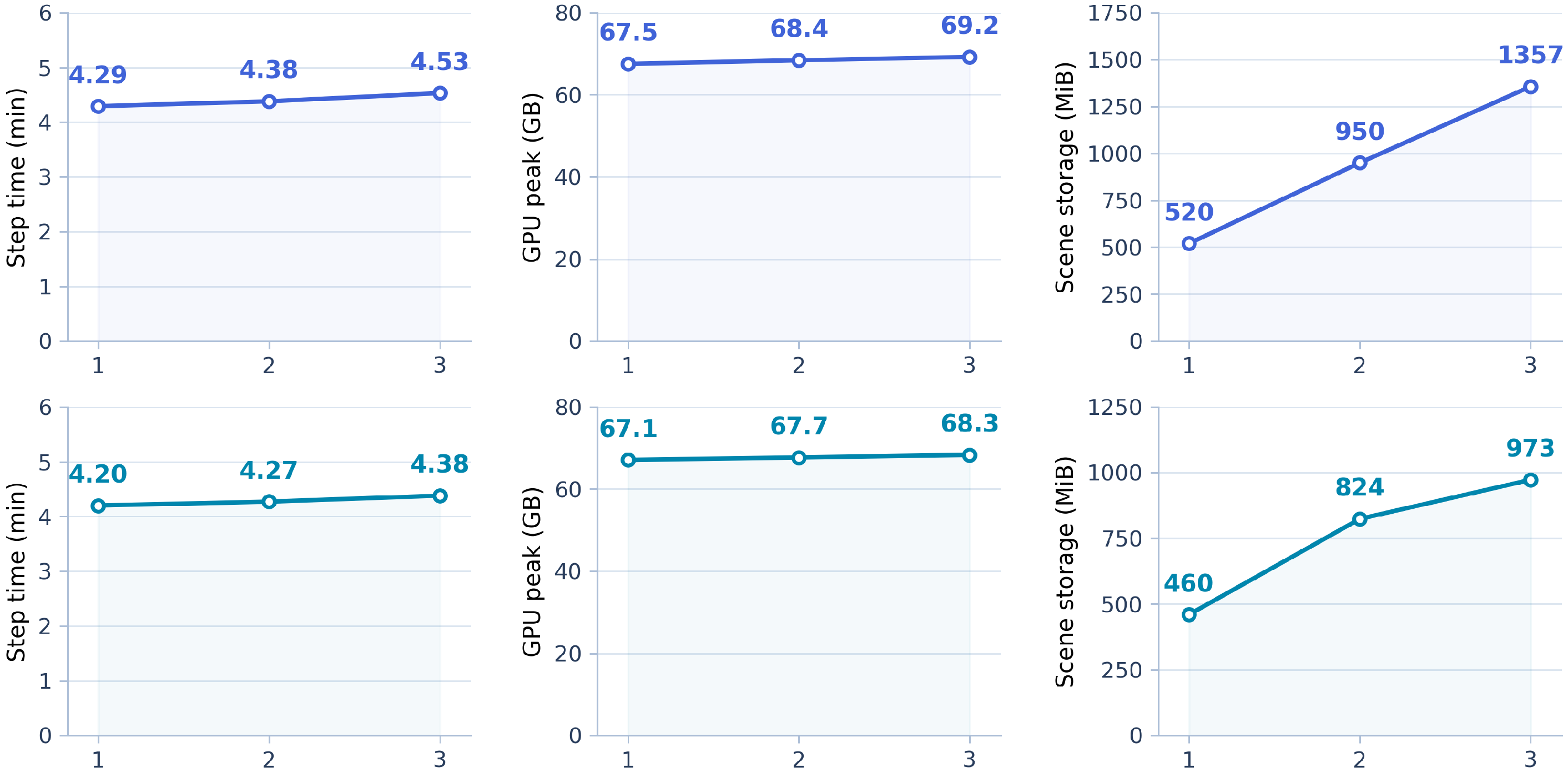}
    \vspace{-10pt}
    \caption{\textbf{Cost across generation steps on WorldScore.}
    Step time, peak GPU memory, and scene storage over three extrapolation
    steps for outdoor (top) and indoor (bottom) scenes.
    Each additional step increases runtime by only a few seconds and
    scene storage by a few hundred MiB, keeping incremental costs modest.}
    \vspace{-3pt}
    \label{fig:cost_scaling}
\end{figure}

\begin{table}[!t]
    \vspace{-3.26pt}
    \centering
    \caption{\textbf{Runtime breakdown of one generation cycle on WorldScore.} World update accounts for only 10.8\%.}
    \label{tab:runtime_breakdown}
    \vspace{-3pt}
    \fontsize{8.5}{9.5}\selectfont
    \renewcommand{\arraystretch}{1}
    \setlength{\tabcolsep}{4.5pt}

    \begin{tabular}{lcccc}
        \toprule
        & \makecell{\textbf{Condition}\\\textbf{rendering}}
        & \makecell{\textbf{Video}\\\textbf{generation}}
        & \makecell{\textbf{World}\\\textbf{update}}
        & \textbf{Total} \\
        \midrule
        Time (s)   & 11.6 & 211.8 & 27.0 & 250.4 \\
        Proportion & 4.6\%  & 84.6\%  & 10.8\% & 100\% \\
        \bottomrule
    \end{tabular}
    \vspace{-7pt}
    \vspace{3.26pt}
\end{table}

\begin{table*}[!t]
    \centering
    \caption{\textbf{Ablation of conditioning modalities and pose injection.} All conditioning modalities contribute to overall performance. ``Joint pose injection'' combines pose with the rendered conditions, while our full model uses a separate RRE attention branch.}
    \label{tab:condition_ablation}
    \fontsize{8.5}{9.5}\selectfont 
    \renewcommand{\arraystretch}{1} 
    \setlength{\tabcolsep}{4.5pt} 

    \begin{tabular}{l|cccccccc}
        \toprule
        \textbf{Method} & \makecell{\textbf{3D}\\\textbf{Consist.}} & \makecell{\textbf{Photo}\\\textbf{Consist.}} & \makecell{\textbf{Style}\\\textbf{Consist.}} & \makecell{\textbf{Camera}\\\textbf{Ctrl.}} & \makecell{\textbf{Object}\\\textbf{Ctrl.}} & \makecell{\textbf{Content}\\\textbf{Align.}} & \makecell{\textbf{Subjective}\\\textbf{Qual.}} & \textbf{Avg.} \\
        \midrule
        Joint pose injection & 88.14 & 84.51 & 93.80 & 96.03 & 65.84 & 73.70 & 65.56 & 81.08 \\
        w/o pose & 88.64 & 85.74 & 93.68 & 95.58 & 65.72 & 73.41 & 65.75 & 81.22 \\
        w/o depth & 87.13 & 88.02 & 93.84 & 97.65 & 65.53 & 74.30 & 66.20 & 81.81 \\
        w/o opacity & 90.06 & 88.91 & 94.06 & 98.82 & 64.34 & 73.06 & 65.87 & 82.16 \\
        w/o pose \& opacity & 87.66 & 85.37 & 93.80 & 95.02 & 64.18 & 72.59 & 65.30 & 80.56 \\
        \midrule
        \textbf{Ours} & \textbf{90.83} & \textbf{88.47} & \textbf{94.40} & \textbf{99.07} & \textbf{65.91} & \textbf{74.70} & \textbf{66.59} & \textbf{82.85} \\
        \bottomrule
    \end{tabular}

\end{table*}

\begin{table*}[!t]
    \centering
    \caption{\textbf{Ablation of persistent world-state update.} Content and geometric filtering, together with cross-chunk alignment, improve multi-step extrapolation and revisit consistency.}
    \label{tab:world_state_update_ablation}

    \fontsize{8}{9}\selectfont
    \renewcommand{\arraystretch}{1}
    \setlength{\tabcolsep}{3.5pt}

    \begin{tabular}{l|cccccccc|cccc}
        \toprule
        \multirow{2}{*}{\textbf{Variant}}
        & \multicolumn{8}{c|}{\textbf{WorldScore Three-step Extrapolation}}
        & \multicolumn{4}{c}{\textbf{Revisit Evaluation}} \\
        \cmidrule(lr){2-9}\cmidrule(lr){10-13}
        & \makecell{\textbf{3D}\\\textbf{Consist.}}
        & \makecell{\textbf{Photo}\\\textbf{Consist.}}
        & \makecell{\textbf{Style}\\\textbf{Consist.}}
        & \makecell{\textbf{Camera}\\\textbf{Ctrl.}}
        & \makecell{\textbf{Object}\\\textbf{Ctrl.}}
        & \makecell{\textbf{Content}\\\textbf{Align.}}
        & \makecell{\textbf{Subjective}\\\textbf{Qual.}}
        & \textbf{Avg.}
        & \textbf{PSNR}$\uparrow$
        & \textbf{SSIM}$\uparrow$
        & \textbf{LPIPS}$\downarrow$
        & \makecell{\textbf{Match}\\\textbf{Ratio}$\uparrow$} \\
        \midrule
        w/o Both Masks
        & 72.78 & 73.97 & 78.16 & 68.15 & 56.09 & 52.58 & 54.23 & 65.14
        & 17.15 & 0.508 & 0.472 & 0.378 \\
        w/o Content Mask
        & 76.05 & 80.30 & 79.86 & 80.15 & 55.53 & 52.11 & 56.80 & 68.69
        & 18.37 & 0.566 & 0.429 & 0.425 \\
        w/o Geometric Mask
        & 74.33 & 81.69 & 82.07 & 76.92 & 56.12 & 52.44 & 57.26 & 68.69
        & 17.65 & 0.537 & 0.433 & 0.447 \\
        w/o Alignment
        & 75.54 & 81.16 & 81.43 & 78.50 & 55.69 & 51.79 & 58.29 & 68.91
        & 17.41 & 0.509 & 0.498 & 0.406 \\
        \midrule
        \textbf{Ours}
        & \textbf{77.50}
        & \textbf{84.51}
        & \textbf{84.36}
        & \textbf{85.17}
        & \textbf{56.25}
        & \textbf{52.82}
        & \textbf{60.34}
        & \textbf{71.56}
        & \textbf{19.85}
        & \textbf{0.623}
        & \textbf{0.369}
        & \textbf{0.481} \\
        \bottomrule
    \end{tabular}
    \vspace{-5pt}
\end{table*}

\subsection{Main Results}



\paragraph{World generation on WorldScore.}
Tables~\ref{tab:worldscore_single_step} and~\ref{tab:worldscore_three_step} report single-step and three-step extrapolation results, respectively. In the single-step setting, GSExplorer achieves the highest average score of \textbf{82.85}, outperforming the strongest baseline by 6.24 points. It reaches \textbf{99.07} Camera Control and also ranks first in 3D Consistency, Style Consistency, Object Control, and Subjective Quality. These results demonstrate precise local control together with coherent geometry and high visual quality.


Under three-step extrapolation, GSExplorer achieves the best average score of \textbf{71.56}, exceeding the strongest baseline, Gen3C, by 7.13 points. It ranks first in 3D Consistency, Style Consistency, and Object Control, and second in Photometric Consistency and Camera Control. WonderWorld's higher Camera Control score benefits from lifting generated images into 3D at prescribed poses and rendering along the given trajectory. The larger margin than in single-step generation suggests that a persistent world state helps sustain performance across chunks. Fig.~\ref{fig:main-generation} further shows accurate camera following and coherent scene structure.

\paragraph{Long-horizon revisit consistency.}
As shown in Table~\ref{tab:revisit_eval}, GSExplorer achieves the best performance on all four revisit metrics, with \textbf{19.85} PSNR, \textbf{0.623} SSIM, \textbf{0.369} LPIPS, and \textbf{0.481} Match Ratio. Relative to the strongest competing results, PSNR improves by 1.99~dB and Match Ratio by 0.050. Figure~\ref{fig:main-revisit} further shows better preservation of the starting-view content and fine details after long-range exploration. These results support the value of a persistent 3D world state in providing reliable conditions when revisiting previously explored regions.

\paragraph{Computational cost.}
Maintaining the 3DGS world state introduces modest overhead. In the WorldScore example in Table~\ref{tab:runtime_breakdown}, world update takes 10.8\% of the generation cycle; including condition rendering brings this share to 15.4\%. Across three steps in Fig.~\ref{fig:cost_scaling}, per-step runtime increases by only 4.2--9.0~s between steps, with small increases in peak GPU memory and a few hundred MiB of additional scene storage. Together with the consistency gains, these costs offer a favorable trade-off for continued exploration. The Appendix provides detailed comparisons with other geometry-conditioned video generation methods, among which GSExplorer achieves the shortest runtime.

\subsection{Ablation Study}

\paragraph{Conditioning design.}
Table~\ref{tab:condition_ablation} shows that each evaluated conditioning modality contributes to overall performance. Removing depth reduces 3D Consistency from 90.83 to 87.13, while removing opacity lowers the average score by 0.69 points. Pose conditioning is important for controllability: removing it reduces Camera Control from 99.07 to 95.58. Joint pose injection retains pose information but combines it with the rendered conditions, underperforming the separate RRE branch in both Camera Control and average score. Removing both pose and opacity causes the largest overall drop, supporting their complementary roles. The full design achieves the best overall performance.

\paragraph{Persistent world-state update.}
Table~\ref{tab:world_state_update_ablation} evaluates content and geometric filtering and cross-chunk alignment. Removing either mask reduces the three-step average from 71.56 to 68.69, while removing both lowers it further to 65.14. Both masks also improve revisit consistency: removing the geometric mask causes a larger PSNR drop, whereas removing the content mask affects Match Ratio more. Without alignment, the average falls to 68.91, with revisit PSNR and Match Ratio decreasing to 17.41 and 0.406. These results show that content filtering, geometric filtering, and alignment complement each other in maintaining a reliable persistent world state.

\section{Conclusion}
GSExplorer shows that interactive world generation can benefit from separating long-term spatial state from local video synthesis. We maintain a persistent 3DGS world state that is repeatedly queried and updated through inconsistency filtering and support-guided cross-chunk alignment. This design keeps generation local while allowing spatial information to accumulate in a reusable scene representation. Results on multi-step extrapolation and long-horizon revisit demonstrate that reliable world-state maintenance improves controllability and consistency with modest update costs. A natural next step is to make world updates more robust to dynamic content and ambiguous observations, while further improving the efficiency and scalability of persistent scene maintenance and generation.

\section*{LLM Usage Statement}
OpenAI's ChatGPT and Codex assisted with manuscript revision, language polishing, related-work organization, code modification and debugging, and plotting scripts. Qwen3.5 generated scene descriptions for training clips. The authors developed the research ideas and experimental designs and made all methodological decisions. All results come from actual experiments, quantitative plots reflect the corresponding measurements, and qualitative examples are outputs of the evaluated models. The authors verified all AI-assisted text and code and take full responsibility for the scientific content, results, and conclusions.
{\small
\bibliographystyle{ieeenat_fullname}
\bibliography{references}
}
\clearpage

\maketitlesupplementary
\appendix
\section*{Appendix}
We provide additional implementation details of the persistent 3DGS world update and the condition-driven video generator in Appendix~A, followed by details of the training-pair construction pipeline in Appendix~B. Appendix~C presents computational cost comparisons and additional qualitative results, while Appendix~D discusses the current limitations of GSExplorer. The supplementary material also includes an MP4 video with extended multi-step exploration and long-horizon revisit examples.

\section{Additional Implementation Details}
\label{app:implementation}

\subsection{Persistent 3DGS World Update}
\label{app:world_update}

\paragraph{3DGS-SLAM backend.}
3DGS-SLAM systems~\citep{yu2025opengs-slam,cheng2025outdoor-s3po-gs} support the viewpoint-dependent rendering and incremental updates required by our world state. They maintain a renderable Gaussian scene~\citep{zhao2026mmgs,zhao2026brpo} while estimating and refining camera states from streaming observations~\citep{cheng2025reggs,yu2026robust-KiloGS-SLAM}. Our ARTDECO-based~\citep{li2025artdeco} 3DGS backend uses MASt3R~\citep{leroy2024grounding-mast3r} correspondences and a Depth Anything 3 (DA3)~\citep{lin2025da3} geometry prior.

The Gaussian scene, historical mapping views, camera graph, and retrieval database persist throughout each rollout. Each 49-frame chunk shares its first frame with the preceding chunk and adds 48 new observations. DA3 jointly processes these frames to initialize depth and cameras, aligning the resulting geometry to the existing reconstruction. Inter-frame correspondences and verified historical matches connect current and previous keyframes for camera optimization.

Tracking keyframes are selected with a match-fraction threshold of $0.45$ and a minimum separation of four global frames. Each chunk retains up to 24 ordinary mapping frames. Frames with insufficient geometric support are excluded before mapping.

\paragraph{Appearance and geometric filtering.}
The consistency checks in Sec.~\ref{sec3-1:world_state} use a historical coverage threshold of $\eta=0.95$. Appearance discrepancy is the channel-wise mean absolute RGB difference, with RGB values in $[0,1]$. The content mask retains covered pixels whose discrepancy does not exceed $\tau_a=0.15$.

Geometric filtering reuses SLAM correspondences and reference associations, using at most $K=3$ reference frames per query. Using pointmaps from the DA3-based mapping pipeline, we compute reprojection errors in coordinates normalized by camera intrinsics. The geometric mask retains pixels whose median error over valid matches is at most $\tau_g=0.01$. Valid matches require finite geometry, positive depth, and projections within the reference image, without additional depth-based occlusion tests. Pixels without valid matches remain unfiltered. The limited reference set avoids traversing the full history for each check.

\begin{table*}[!t]
    \centering
    \caption{\textbf{Conditioning architectures and added parameters of geometry-conditioned video models.} Ratios are computed from unrounded parameter counts relative to the nominal backbone sizes. GSExplorer introduces 0.19B parameters, equivalent to 3.8\% of its backbone. ``--'' denotes unavailable parameter counts.}
    \label{tab:conditioning_architecture}
    \vspace{-3pt}
    \fontsize{8.5}{9.5}\selectfont
    \renewcommand{\arraystretch}{1}
    \setlength{\tabcolsep}{4.5pt}

    \begin{tabular}{llccl}
        \toprule
        \textbf{Method}
        & \textbf{Video Backbone}
        & \textbf{Added Params. (B)}
        & \textbf{Backbone Ratio}
        & \textbf{Condition Injection} \\
        \midrule
        Gen3C~\citep{ren2025gen3c}
        & Cosmos Predict1-7B
        & -- & --
        & Channel-wise concatenation \\
        Uni3C~\citep{cao2025uni3c}
        & Wan2.1-I2V-14B
        & 1.00 & 7.1\%
        & ControlNet branch \\
        Voyager~\citep{huang2025voyager}
        & HunyuanVideo-I2V-13B
        & 0.50 & 3.9\%
        & Context branch + residual injection \\
        NeoVerse~\citep{yang2026neoverse}
        & Wan2.1-T2V-14B
        & 3.05 & 21.8\%
        & Control branch \\
        Spatia~\citep{zhao2026spatia}
        & Wan2.2-TI2V-5B
        & 1.48 & 29.5\%
        & VACE branch + LoRA \\
        UniWorld-View~\citep{zhou2026uniworld-view}
        & Wan2.1-VACE-14B
        & 3.47 & 24.8\%
        & VACE + reference-feature cross-attention \\
        PPMem~\citep{he2026ppmem}
        & Wan2.2-TI2V-5B
        & -- & --
        & Channel-wise concatenation \\
        \midrule
        \textbf{GSExplorer (Ours)}
        & Wan2.2-TI2V-5B
        & \textbf{0.19} & \textbf{3.8\%}
        & Geometry adapter + pose RRE \\
        \bottomrule
    \end{tabular}
    \vspace{-5pt}
\end{table*}

\paragraph{Support-guided cross-chunk alignment.}
Camera alignment proceeds in two stages. First, we retain inter-frame and verified loop-closure matches that satisfy the backend's validity and confidence checks and have supported endpoints. These matches constrain cross-chunk bundle adjustment using the backend's native optimization variables and residuals.

Second, we refine current-chunk camera poses using supported correspondences, MASt3R canonical pointmaps, and the backend's native residuals. This step optimizes only current-chunk rotations and translations for at most five Gauss--Newton iterations. Historical poses and estimated scales remain fixed at their post-BA values, and pointmaps remain unchanged.

For this refinement, 20\% of correspondence endpoints are withheld from the solve for candidate validation. Candidates must improve the supported-subset objective while increasing the full validation cost by at most 1\% relative to the post-BA poses, a tolerance selected through preliminary small-scale experiments. Pose corrections are bounded and may be reduced or rejected; if none is accepted, the post-BA poses are retained. This holdout applies only to the additional refinement, since these observations may have contributed to BA or mask construction.

Both optimization stages select correspondence endpoints using current-frame masks computed at the initial poses and stored acceptance masks for historical frames. After refinement, masks for accepted current frames are recomputed at the final poses before Gaussian insertion and optimization. Historical masks are retained for reuse.

\paragraph{Selective Gaussian scene update.}
The support mask $S_i$ selects observations for Gaussian insertion and scene optimization, subject to validity checks for finite 3D points, positive depth, and finite RGB values. Newly revealed regions also undergo geometric filtering. Before Gaussian insertion, rejected regions are expanded with a $7\times7$ neighborhood to reduce conflicting supervision near their boundaries. During scene refinement, the support mask is applied to RGB, SSIM, and inverse-depth supervision.

\subsection{Condition-Driven Video Generation}
\label{app:generation_model}

\paragraph{Multimodal condition adapter.}
Rendered RGB and normalized depth are independently encoded by the same frozen Wan VAE, producing 48-channel latent features for each modality. Accumulated opacity and the binary hole mask are concatenated and processed by a lightweight 3D convolutional encoder with 64 output channels. These features are concatenated into 160 channels and projected to the 3072-dimensional DiT hidden space.

The projected condition is injected at the DiT input and after blocks $\{5,10,15,20,25\}$, using zero-based indexing. All six conditioning projections are zero-initialized, preserving the pretrained backbone's initial behavior while allowing it to gradually learn the additional scene conditions.


\paragraph{Relative Ray Encoding.}
We use the Relative Ray Encoding (RRE) of UCPE~\citep{zhang2026unified-UCPE}, without Absolute Orientation Encoding. Camera intrinsics and camera-to-world poses define each latent token's local viewing-ray frame. Each RRE module projects the 3072-dimensional hidden state to three 128-dimensional attention heads. Queries, keys, and values undergo geometric transforms in their ray frames, followed by the inverse transform on the attention output.

The RRE branch operates in parallel with the original self-attention in every DiT block. Its output is projected back to 3072 dimensions through a zero-initialized projection and added to the original attention output before the subsequent cross-attention and feed-forward operations.

Table~\ref{tab:conditioning_architecture} compares conditioning architectures and parameter overhead across geometry-conditioned video models. Our multimodal adapter and RRE branch together add only 0.19B parameters, or 3.8\% of the backbone size. This lightweight design supports strong camera control and generation quality, as shown in Tables~\ref{tab:worldscore_single_step} and~\ref{tab:worldscore_three_step}.

\paragraph{Geometric normalization.}
We divide depth by the 98th percentile of valid depth over the 49-frame clip and clip it to $[0,1]$. Camera translations relative to the first frame use the same normalization scale. Valid pixels require accumulated opacity of at least $0.95$ and finite depth within the configured valid range. The binary hole mask is:
\begin{equation}
    M_{\mathrm{hole}}(u)=\mathbf{1}[A(u)<0.95],
\end{equation}
while accumulated opacity remains a continuous condition.

\paragraph{Training details.}
Following Sec.~4.1, we use BF16, AdamW with zero weight decay, and cosine learning-rate schedules. For the first 10K iterations, the DiT is frozen while the multimodal adapter and RRE branch use a peak learning rate of $10^{-4}$. For the remaining 20K iterations, peak rates are $10^{-7}$ for the DiT and $10^{-5}$ for both branches. The VAE and text encoder remain frozen throughout.

We apply clean-reference dropout with probability $0.5$. When the reference is retained, its clean latent replaces the first noisy temporal slice, which is excluded from the flow-matching loss as in Eq.~\ref{eq:flow_matching}. When the reference is dropped, all temporal slices are noised and contribute to the loss. The reference supports cross-chunk continuity, while dropout encourages the model to use the scene conditions. Text prompts are dropped independently with probability $0.1$.

For the weighted flow-matching objective in Sec.~\ref{3.2-generation model}, the training scheduler uses $L=1000$ noise levels with shift $5$. At scheduler noise level $\sigma_k$, the weight is:
\begin{equation}
    \begin{aligned}
        a_k
        &= \exp\!\left[-2(\sigma_k-\tfrac12)^2\right], \\
        w_k
        &= \frac{a_k-a_{\min}}
        {\frac{1}{L}\sum_{j=1}^{L}(a_j-a_{\min})},
    \end{aligned}
    \label{eq:noise_weights}
\end{equation}
where $a_{\min}=\min_{1\leq j\leq L}a_j$ and $w(\sigma_k)=w_k$. This normalizes the weights to unit mean over the scheduler noise levels. No additional geometric, reprojection, perceptual, temporal-consistency, or RGB reconstruction losses are used for generator training.

\begin{table*}[!t]
    \centering
    \caption{\textbf{Computational cost comparison on WorldScore.} GSExplorer achieves the shortest runtime among the compared geometry-conditioned video generation methods. Its relatively high but manageable memory usage offers a favorable trade-off for stronger generation consistency. All measurements are obtained on a single NVIDIA H20 GPU without offloading.}
    \label{tab:worldscore_cost}
    \vspace{-4pt}
    \fontsize{9}{9.5}\selectfont
    \renewcommand{\arraystretch}{1.05}
    \setlength{\tabcolsep}{5.5pt}

    \begin{tabular}{lc|cc|cc|cc}
        \toprule
        \multirow{2}{*}{\textbf{Method}}
        & \multirow{2}{*}{\makecell{\textbf{Output}\\\textbf{Resolution}}}
        & \multicolumn{2}{c|}{\textbf{Runtime (min)}$\downarrow$}
        & \multicolumn{2}{c|}{\textbf{Peak GPU Memory (GiB)}$\downarrow$}
        & \multicolumn{2}{c}{\textbf{WorldScore Avg.}$\uparrow$} \\
        \cmidrule(lr){3-4}\cmidrule(lr){5-6}\cmidrule(lr){7-8}
        & & \textbf{1-step} & \textbf{3-step}
          & \textbf{1-step} & \textbf{3-step}
          & \textbf{1-step} & \textbf{3-step} \\
        \midrule
        Gen3C~\citep{ren2025gen3c}
        & $1280\times704$
        & 19.1 & 59.2 & 58.9 & 66.2 & 71.74 & 64.43 \\
        Uni3C~\citep{cao2025uni3c}
        & $1280\times704$
        & 21.2 & 65.2 & 71.1 & 71.4 & 70.25 & 56.58 \\
        Voyager~\citep{huang2025voyager}
        & $768\times512$
        & 32.8 & 98.4 & 76.6 & 76.6 & 58.04 & 33.91 \\
        NeoVerse~\citep{yang2026neoverse}
        & $560\times336$
        & 10.8 & 33.5 & 57.4 & 57.5 & 72.31 & 60.11 \\
        Spatia~\citep{zhao2026spatia}
        & $1248\times704$
        & 6.1 & 36.4 & 51.4 & 62.9 & 70.38 & 48.82 \\
        UniWorld-View~\citep{zhou2026uniworld-view}
        & $832\times480$
        & 6.3 & 20.9 & 70.8 & 70.8 & 76.61 & 64.08 \\
        PPMem~\citep{he2026ppmem}
        & $1024\times576$
        & 8.6 & 30.9 & \textbf{38.7} & \textbf{44.5} & 71.69 & 58.86 \\
        \midrule
        \textbf{GSExplorer (Ours)}
        & $1280\times704$
        & \textbf{4.2} & \textbf{12.7}
        & 67.6 & 69.8
        & \textbf{82.85} & \textbf{71.56} \\
        \bottomrule
    \end{tabular}
    \vspace{-6pt}
\end{table*}

\section{Training Pair Construction Details}
\label{app:data}

\subsection{Dataset Statistics}
\label{app:data_statistics}

After preprocessing, reconstruction, and alignment filtering, the final training set contains 293,943 video pairs. Table~\ref{tab:data_statistics} summarizes their composition.

\begin{table}[t]
    \centering
    \caption{Composition of the final training set.}
    \label{tab:data_statistics}
    \vspace{-3pt}

    \fontsize{8.5}{9.5}\selectfont
    \renewcommand{\arraystretch}{1.05}
    \setlength{\tabcolsep}{2.3pt}

    \begin{tabular}{@{}l|ccccc@{}}
        \toprule
        \textbf{Dataset}
        & \textbf{RealEstate10K}
        & \textbf{DL3DV}
        & \textbf{ACID}
        & \textbf{ScanNet}
        & \textbf{Total} \\
        \midrule
        \textbf{Training pairs}
        & 231,644
        & 14,279
        & 9,983
        & 38,037
        & \textbf{293,943} \\
        \bottomrule
    \end{tabular}
    \vspace{-5pt}
\end{table}


\subsection{Data Sampling and Filtering}
\label{app:data_filtering}

All datasets use 57-frame clips containing 8 context frames and 49 target frames, with RGB images, intrinsics, and camera poses sampled at matching indices. DL3DV and ScanNet use consecutive 57-frame windows with a stride of 28. For RealEstate10K, the context frames are consecutive, while target frames are sampled over the remaining valid interval. ACID sequences are sampled approximately uniformly over temporally and geometrically valid intervals.

We screen candidate clips using bidirectional depth-reprojection coverage and relative camera rotation. Given directional coverage $c_{i\rightarrow j}$, the symmetric overlap is:
\begin{equation}
    o_{ij}
    =
    \min(c_{i\rightarrow j},c_{j\rightarrow i}).
\end{equation}
DL3DV and ScanNet retain anchor views with overlap of at least $0.35$ and relative rotation below $50^\circ$. RealEstate10K and ACID additionally use DA3-estimated geometry to truncate or split sequences with excessive viewpoint changes.

We reject clips with invalid cameras, severe image degradation, temporal discontinuities, or dominant dynamic content. Additional checks include DOVER-based quality filtering for DL3DV and ScanNet, transition detection for RealEstate10K, and checks for persistent borders, exposure changes, and highly dynamic foreground regions for ACID.

\subsection{Pair Construction and Alignment Filtering}
\label{app:pair_construction}

Following Sec.~\ref{sec3-3:training_pairs}, we reconstruct all 57 frames and force the first target frame (zero-based index 8) to be a keyframe for Gaussian insertion. We record each Gaussian's birth frame $b_g$ and retain context primitives satisfying:
\begin{equation}
    0\le b_g<8.
\end{equation}

The retained primitives keep their parameters from full-sequence optimization. This preserves the improved camera and geometric alignment while removing Gaussians introduced by target frames.

We render the context map at the 49 estimated target poses to obtain RGB, depth, and accumulated opacity. Translation-only ECC estimates alignment between the final rendered frame and its real target over a valid region determined from opacity and conservatively eroded before registration. Clips with estimated translation magnitude above $4$ pixels are discarded. ECC only verifies alignment; its transformation is not applied to the rendered conditions.

\section{Additional Results}
\label{app:qualitative}

\paragraph{Computational cost.}
Table~\ref{tab:worldscore_cost} compares runtime, GPU memory usage, and generation performance on WorldScore. GSExplorer achieves the shortest runtime among the compared geometry-conditioned video generation methods in both single-step and three-step settings, while attaining the highest overall scores. Its memory usage is relatively high but remains manageable, offering a favorable trade-off for stronger generation consistency.

\begin{figure*}[t]
    \centering
    \includegraphics[width=\linewidth]{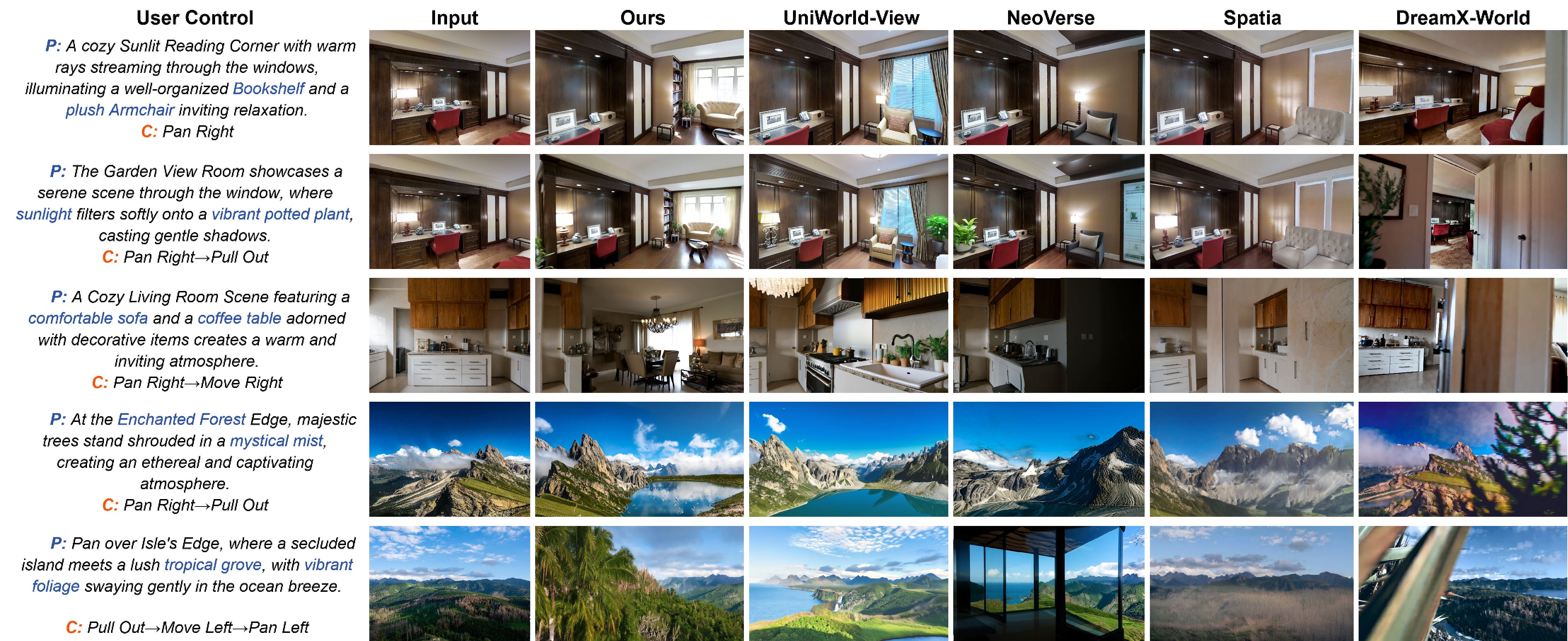}
    \caption{
    Additional qualitative results on WorldScore.
    GSExplorer follows the specified text and camera controls while maintaining coherent scene appearance and geometry during extrapolation.
    }
    \label{fig:appendix-generation}
\end{figure*}

\begin{figure*}[t]
    \centering
    \includegraphics[width=\linewidth]{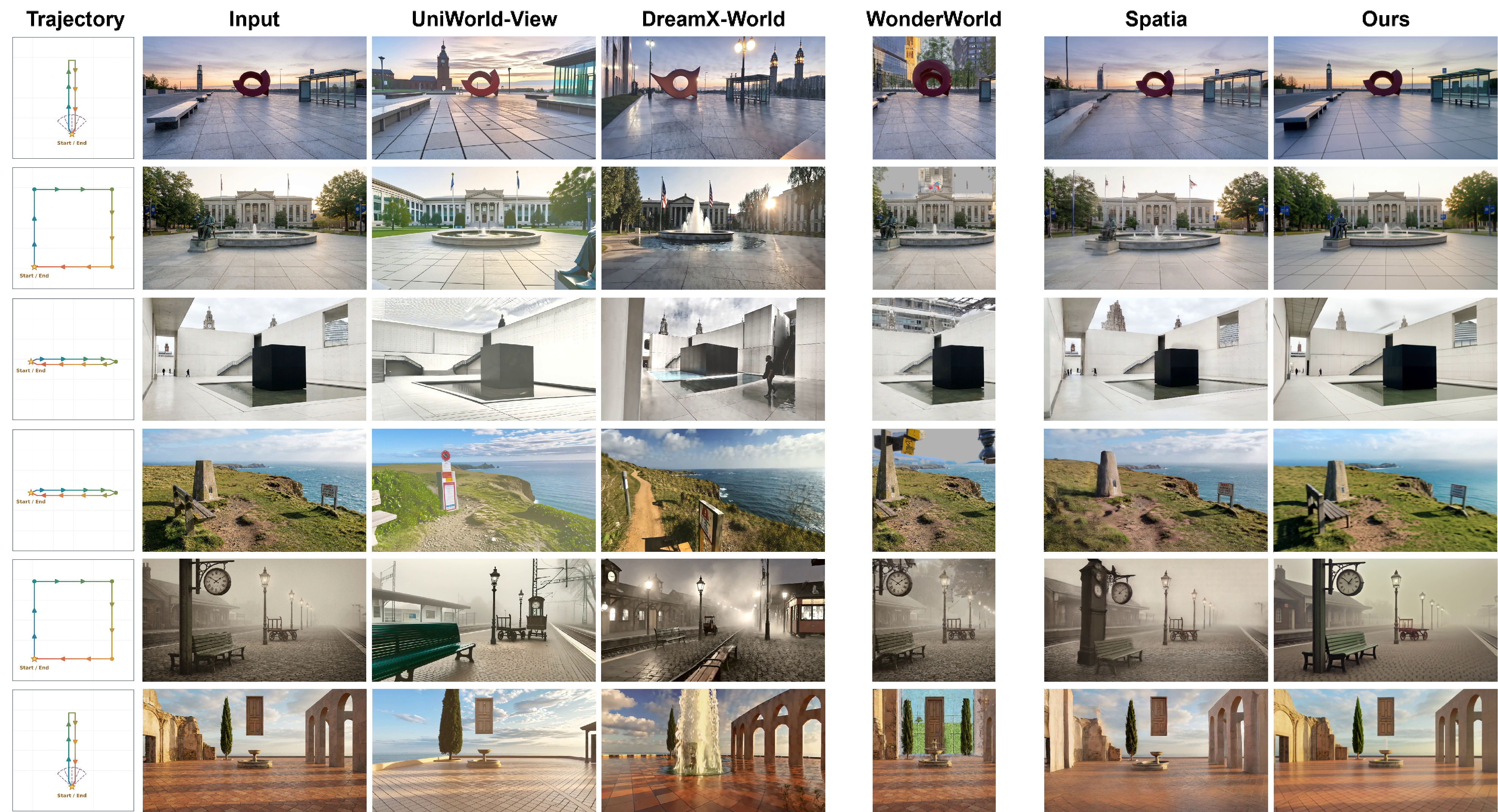}

    \caption{
    Additional qualitative comparison of long-horizon revisit results.
    After extended leave-and-return exploration, GSExplorer better preserves the scene content and spatial layout of the initial view.
    }
    \label{fig:appendix-revisit}
\end{figure*}

\paragraph{Qualitative results.}
Figure~\ref{fig:appendix-generation} provides additional examples of text- and camera-controlled world generation on WorldScore. Figure~\ref{fig:appendix-revisit} presents additional long-horizon leave-and-return comparisons. More complete rollouts are included in the supplementary MP4 video.

\section{Limitations}
\label{app:limitations}

Although maintaining the 3DGS world state adds modest overhead, runtime is dominated by the current bidirectional video generator. GSExplorer is also primarily designed for static or predominantly static environments, and its world state does not explicitly model dynamic objects. Future work will focus on accelerating generation while preserving its quality and extending the world state to dynamic scenes, enabling faster exploration and richer interactions.

\end{document}